\documentclass[letterpaper, 10 pt, conference]{ieeeconf}  % Comment this line out if you need a4paper
\usepackage{amsmath}
\usepackage{graphicx}
\usepackage{caption}
\usepackage{booktabs}
\usepackage{amssymb}
\usepackage{bm}
\usepackage[colorlinks,linkcolor=black, citecolor=black, urlcolor=black]{hyperref}
\usepackage{algorithmicx}
\usepackage[noend]{algpseudocode}
\usepackage{algorithm}
\usepackage{cite}
\usepackage{subfig}
\usepackage{url}
\usepackage[nameinlink,capitalise]{cleveref}
\usepackage{multirow}
\usepackage{xcolor}
\newcommand{\AND}{\text{and}}
\usepackage{balance}
\usepackage{setspace}
\IEEEoverridecommandlockouts                              % This command is only needed if
\title{\LARGE \bf
Fast Safe Rectangular Corridor-based Online AGV Trajectory Optimization with Obstacle Avoidance
}
\author{Shaoqiang Liang$^{1}$, Songyuan Fa$^{1}$ and Yiqun Li$^{1}$% <-this % stops a space
\thanks{This research was funded by the National Natural Science Foundation of China (Grant No.~51905185) and the National Postdoctoral Program for Innovative Talents No. BX20180109.(\textit{Corresponding author: Yiqun Li)}.}% <-this % stops a space
\thanks{$^{1}$Shaoqiang Liang, Songyuan Fa and Yiqun Li are affiliated with the State Key Laboratory of Intelligent Manufacturing Equipment and Technology, School of Mechanical Science and Engineering, Huazhong University of Science and Technology, Wuhan 430074, China.
       {\tt\small sqliang@hust.edu.cn, liyiqun@hust.edu.cn}}
}

\begin{document}
\maketitle
\thispagestyle{empty}
\pagestyle{empty}

%%%%%%%%%%%%%%%%%%%%%%%%%%%%%%%%%%%%%%%%%%%%%%%%%%%%%%%%%%%%%%%%%%%%%%%%%%%%%%%%
\begin{abstract}
Automated Guided Vehicles (AGVs) are essential in various industries for their efficiency and adaptability. However, planning trajectories for AGVs in obstacle-dense, unstructured environments presents significant challenges due to the nonholonomic kinematics, abundant obstacles, and the scenario's nonconvex and constrained nature. To address this, we propose an efficient trajectory planning framework for AGVs by formulating the problem as an optimal control problem. Our framework utilizes the fast safe rectangular corridor (FSRC) algorithm to construct rectangular convex corridors, representing avoidance constraints as box constraints. This eliminates redundant obstacle influences and accelerates the solution speed. Additionally, we employ the Modified Visibility Graph algorithm to speed up path planning and a boundary discretization strategy to expedite FSRC construction. Experimental results demonstrate the effectiveness and superiority of our framework, particularly in computational efficiency. Compared to advanced frameworks, our framework achieves computational efficiency gains of 1 to 2 orders of magnitude. Notably, FSRC significantly outperforms other safe convex corridor-based methods regarding computational efficiency.
\end{abstract}
\setlength{\abovedisplayskip}{1.2pt}
\setlength{\belowdisplayskip}{1.2pt}
%%%%%%%%%%%%%%%%%%%%%%%%%%%%%%%%%%%%%%%%%%%%%%%%%%%%%%%%%%%%%%%%%%%%%%%%%%%%%%%%
\section{INTRODUCTION}
Recently, the deployment of Automated Guided Vehicles (AGVs) has witnessed a significant upsurge across diverse industrial and logistical contexts \cite{saleh2019real}. Their growing popularity can be attributed to their efficiency and adaptability \cite{li2019real}, finding applications in sectors such as mining \cite{roberts2002reactive}, surveillance \cite{zhang2022agvs}, forklift operations \cite{li2015vision}, traffic management \cite{pratissoli2021hierarchical}, electronic manufacturing \cite{zou2021effective}, delivery \cite{chen2021integrated} and IoT scenarios \cite{elsisi2021development}. AGVs promise to automate warehouse operations, curb operational costs, and enhance overall production processes \cite{bechtsis2017sustainable, vlachos2023lean}. However, effectively employing AGVs in obstacle-rich environments \cite{farooq2021flow, mercy2017spline} remains a considerable challenge.

Trajectory planning plays a crucial role in the functioning of AGV systems. As AGVs traverse intricate and ever-evolving environments, their primary task is to navigate while avoiding collisions with stationary and moving obstacles. The challenges of real-time AGV operations underscore the critical role played by efficient trajectory planning frameworks. In obstacle-rich, unstructured environments, trajectory planning for AGVs remains daunting, primarily due to their nonholonomic kinematics, the abundance of obstacles, and the nonconvex, constrained nature of the scenario. This paper aims to develop efficient and optimal trajectory planning solutions for AGVs operating in such environments.
\subsection{Related Works}
Optimization-based methods describe trajectory planning as an optimal control problem (OCP) \cite{wang2023autonomous,li2021fast,li2016time,shi2019bilevel}, solving an optimization problem to obtain the optimal trajectory. This approach excels at finding trajectories that meet multiple constraints simultaneously. Thus, this paper adopts this method to address AGV trajectory planning, aiming to obtain collision-free and kinematically feasible trajectories. However, solving optimization problems in this manner is demanding and time-consuming \cite{li2016time}, and it may become intractable in cases of excessively complex constraints.
\begin{figure}[htbp]
	\centering
	\includegraphics[width=8.4cm]{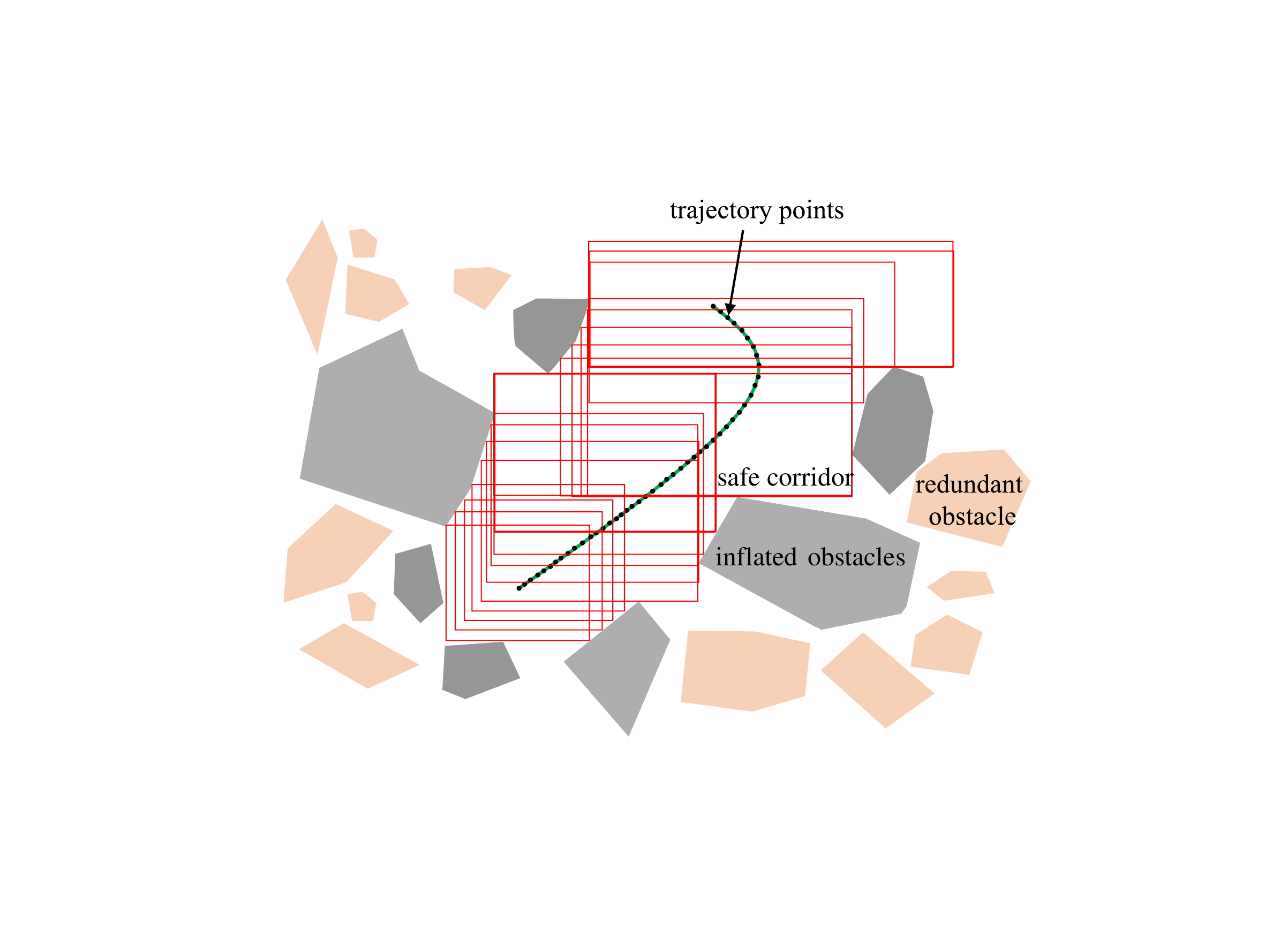}
	\caption{
		Illustration of safe convex corridor-based methods: the red rectangles represent the constructed convex corridors, with each trajectory point corresponding to a rectangle. The light orange areas indicate redundant obstacles.}
	\label{fig:SafeCorridorIllustration}
\end{figure}

Bergman et al. \cite{bergman2018combining} proposed a systematic approach using numerical optimal control techniques to compute local solutions for motion planning. Their method combines Sequential Quadratic Programming with homotopy methods to transform the problem into its original form, enabling the computation of locally optimal solutions for motion planning. Chai et al. \cite{chai2018two} introduced a two-stage trajectory optimization framework for generating optimal parking trajectories for vehicles. Their approach employs a multi-level optimization strategy to enhance convergence and computation efficiency. However, their methods struggle to maintain real-time performance in obstacle-dense scenarios. Guo et al. \cite{guo2022down} proposed a reduced initialization strategy for optimization-based trajectory planning.

Unstructured environments are characterized by irregularly shaped, non-convex obstacles distributed irregularly and in large numbers. As a result, trajectory optimization problems in these environments typically involve numerous collision avoidance constraints, which can decrease solution efficiency. In contrast, safe convex corridor-based methods are better suited for addressing such complex scenarios \cite{li2020maneuver,li2023trajectory,lian2023trajectory,zhu2015convex,liu2017planning}. These methods transform general nonlinear, non-convex collision avoidance constraints into convex constraints by confining the robot within a convex polygon or polyhedron, thereby reducing redundant collision avoidance constraints and easing the optimization problem's difficulty (see Fig.~\ref{fig:SafeCorridorIllustration}).

Deits et al. \cite{deits2015computing} introduced the IRIS (Iterative Regional Inflation by Semidefinite Programming) algorithm, which efficiently computes large polytopes and ellipsoidal free spaces using a series of convex optimizations. Chen et al. \cite{chen2016online} proposed the safe passage algorithm to generate collision-free trajectories in cluttered environments. This approach uses rectangles to construct convex corridors instead of convex polygons as in IRIS. Liu et al. \cite{liu2018convex} presented the Convex Feasible Set Algorithm (CFS), transforming the original problem into a series of convex subproblems. This is achieved by obtaining convex feasible sets within non-convex domains and iteratively solving these subproblems with convex constraints until convergence.
Additionally, Li et al. \cite{li2020maneuver} introduced Safe Travel Corridors (STC) for the trajectory planning of autonomous vehicles. STC simplifies collision avoidance constraints in automated parking motion planning, improving efficiency in complex environments. The construction time of safe convex corridors significantly impacts the overall efficiency of trajectory planning frameworks, thereby affecting real-time performance. However, some algorithms exhibit prolonged execution times, consequently diminishing the planning efficiency.

Additionally, the efficiency of optimization-based methods is influenced by the quality of the initial solution \cite{li2022online}, highlighting the importance of choosing the appropriate path planning algorithm to generate the initial solution. However, algorithms like A* \cite{hart1968formal} and hybrid A* \cite{buehler2009junior} are constrained by map size and precision, while RRT \cite{lavalle2001randomized} and RRT* \cite{noreen2016optimal} exhibit instability and do not guarantee the shortest path.
\subsection{Motivations, Contributions, and Organization}
This study aims to find optimal collision-free trajectories quickly for AGVs in cluttered and unstructured environments. To achieve this, the trajectory planning problem is formulated as an optimal control problem and then solved numerically to obtain trajectories that satisfy various constraints. However, in unstructured environments, solving this problem efficiently is challenging due to the abundance of obstacles. In many studies, including \cite{chen2015decoupled,li2015unified}, excluding redundant collision avoidance constraints remains difficult. These constraints, being nonlinear and nonconvex, often result in inefficient solutions. This paper proposes the FSRC algorithm to address these challenges, which constructs convex corridors using rectangles to ensure AGV motion safety within specific rectangular areas. Additionally, we optimize the entire trajectory planning framework to enhance efficiency and enable real-time motion planning. The specific contributions of this paper are as follows:

\begin{enumerate}
\item
The FSRC algorithm is introduced, significantly accelerating the process of constructing convex regions compared to STC \cite{li2020maneuver} and SFC \cite{li2020maneuver}. The corridors established by FSRC serve as representations for obstacle avoidance constraints within OCP, effectively eliminating redundant obstacles.
\item
The Modified Visibility Graph (MVG) is proposed, demonstrating remarkable path-searching speeds.
\item
A boundary discretization strategy is introduced, enhancing the construction speed of FSRC.
\end{enumerate}

The structure of this paper is as follows: Section 2 defines the discrete trajectory planning problem. Section 3 elaborates on the specific implementation algorithms of the trajectory planning framework, covering obstacle inflation and discretization, the MVG, and the FSRC. Section 4 presents experimental results, and Section 5 concludes the paper.
\section{PROBLEM FORMULATION}
Based on the objective of performing trajectory planning and obstacle avoidance for AGVs, this section introduces an optimal control problem to describe the trajectory planning problem.
\subsection{Formulation of Trajectory Planning Problem}
The workspace for AGV is defined as $W \subset {\mathbb{R}^2}$, with the area occupied by obstacles denoted by ${W_{{\text{obs}}}} \subset W$. Therefore, the AGV must operate in the free space ${W_{{\text{free}}}} \subset W \backslash W_{{\text{obs}}}$. The trajectory planning problem for the AGV aims to find the optimal trajectory connecting the start and end poses of the AGV, satisfying all constraints. Here, the AGV's state profile is represented by $\xi(t) = {[x(t),y(t),v(t),\theta (t)]^T}$ and the control profile by $u(t) = {[a(t),\omega (t)]^T}$. The trajectory planning of the AGV is abstracted as
\begin{gather}
	\label{ex:tpp}
	\begin{aligned}
		& \text{Minimize} \quad J = T_f - T_0 \\
		& \text{Subject to} \\
		& \dot \xi(t) = f\left( {\xi(t),u(t)} \right) \\
		& {{\mathbf{\xi }}_{\min }} \leqslant \xi (t) \leqslant {{\mathbf{\xi }}_{\max }} \\
		& {{\text{U}}_{\min }} \leqslant u(t) \leqslant {{\text{U}}_{\max }}  \\
		&\xi ({T_0}) = {\xi _0},u({T_0}) = {{\text{u}}_0}  \\
		&\xi ({T_f}) = {\xi _f},u({T_f}) = {{\text{u}}_f}  \\
		&g(\xi (t)) \subset {W_{{\text{free}}}},\forall t \in [{T_0},{T_f}],
	\end{aligned}
\end{gather}
where $T_0$ and $T_f$ denote the starting and ending times, $T_f - T_0$ is the cost function, $\dot \xi(t) = f\left( {\xi(t),u(t)} \right) $ describes the AGV kinematics, $ [{\mathbf{\xi }}_{\min }, {\mathbf{\xi }}_{\max }]$ and $[{\text{U}}_{\min }, {\text{U}}_{\max }]$ denote the allowable intervals for $\xi$ and $\text{u}$ respectively, and $g(\xi (t)) \subset {W_{{\text{free}}}}$ denotes collision-avoidance constraints.
\subsection{Discrete Formulation of Trajectory Planning Problem}
Discretizing the interval $[T_0, T_f]$ into $N-1$ segments using the forward Euler method \cite{mathews2004numerical} yields the discrete formulation of the trajectory planning problem, as elaborated in the following equations:
\begin{figure}[htbp]
	\centering
	\includegraphics[width=6cm]{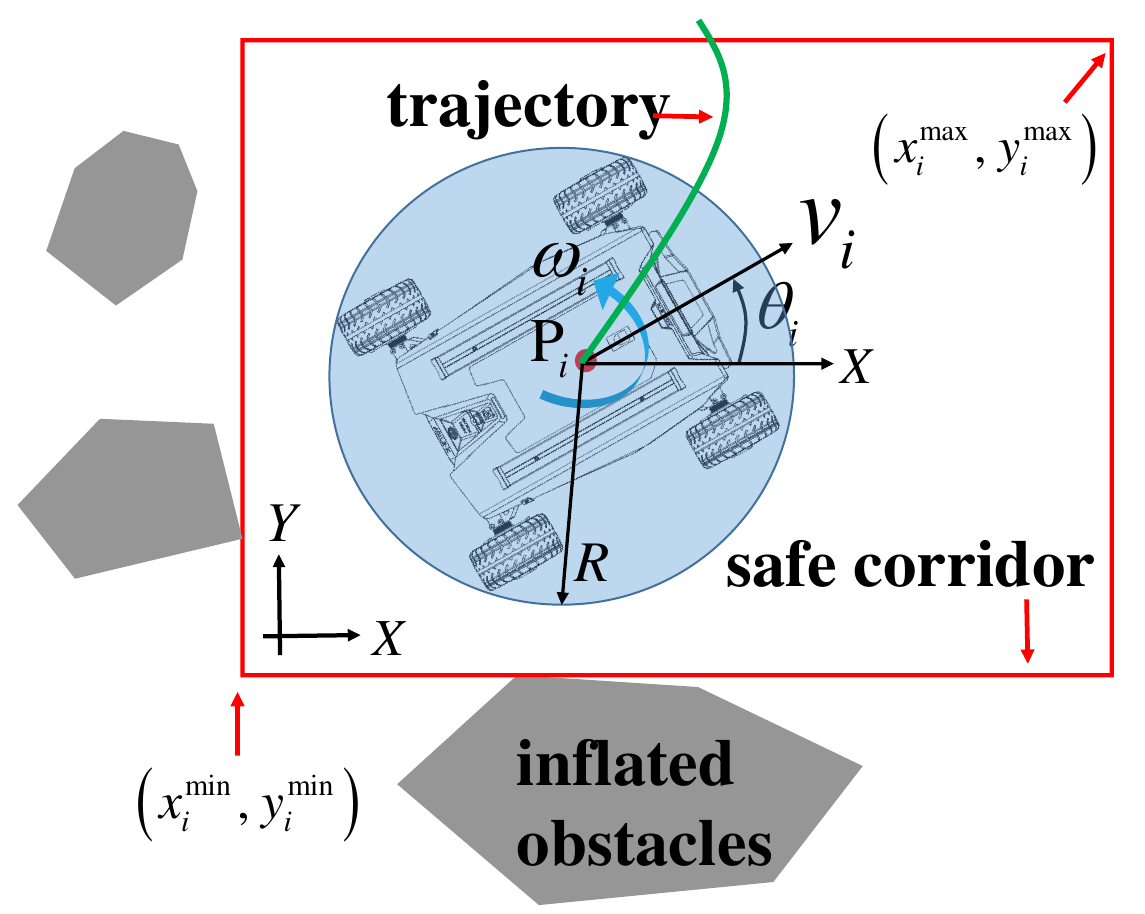}
	\caption{
		The kinematic model of the AGV.}
	\label{fig:scc}
\end{figure}
\begin{gather}
\label{ex:OM}
\begin{aligned}
& \text{Minimize} \quad J = T_f - T_0 \quad (\textnormal{2.a}) \\
& \text{Subject to} \\
& {x_{i + 1}} - {x_i} = v_i \cdot \cos  {{\theta _i}}\cdot {\Delta _{\text{t}}},\ i = 1,2,...,N - 1 \quad (\textnormal{2.b})\\
& {y_{i + 1}} - {y_i} = v_i  \cdot \sin  {{\theta _i}}  \cdot {\Delta _{\text{t}}},\ i = 1,2,...,N - 1 \quad (\textnormal{2.c})\\
& {v_{i + 1}} - {v_i} = a_i \cdot {\Delta _{\text{t}}},\ i = 1,2,...,N - 1 \quad (\textnormal{2.d}) \\
& {\theta_{i + 1}} - {\theta_i} = \omega_i \cdot {\Delta _{\text{t}}},\ i = 1,2,...,N - 1 \quad (\textnormal{2.e}) \\
& x_1 = {x_{{T_0}}},\ y_1 = {y_{{T_0}}},\ \theta_1 = {\theta_{{T_0}}} \quad (\textnormal{2.f}) \\
& v_1 = {v_{{T_0}}},\ a_1 = {a_{{T_0}}},\ \omega_1 = {\omega_{{T_0}}} \quad (\textnormal{2.g}) \\
& x_N = {x_{{T_f}}},\ y_N = {y_{{T_f}}} \quad (\textnormal{2.h}) \\
& v_N = {v_{{T_f}}},\ a_N = {a_{{T_f}}},\ \omega_N = {\omega_{{T_f}}} \quad (\textnormal{2.i}) \\
& x_i^{\min} \leq x_i \leq x_i^{\max},\ y_i^{\min} \leq y_i \leq y_i^{\max},\ i = 1,...,N  \quad (\textnormal{2.j}) \\
& {v}^{\min} \leq v_i \leq {v}^{\max},\ {a}^{\min} \leq a_i \leq {a}^{\max},\ i = 1,...,N  \quad (\textnormal{2.k}) \\
& {\omega}^{\min} \leq \omega_i \leq {\omega}^{\max},\ i = 1,2,...,N , \quad (\textnormal{2.l})
\end{aligned}
\end{gather}
where ${\Delta _t} = \frac{{{T_f} - {T_0}}}{{N - 1}}$ represents the time interval between two time points, equations (2.b$\sim$2.e) capture kinematic constraints for omnidirectional movement. $P_i(x_i, y_i)$ signifies geometric center (see Fig.~\ref{fig:scc}), while $v_i$, $a_i$, $w_i$, and $\theta_i$ respectively represent velocity, acceleration, angular velocity, and yaw angle at $P_i$. (2.f-2.i) define the boundary constraints for the AGV's starting and goal poses. (2.j) represent the obstacle avoidance constraint implemented by FSRC. (2.k) and (2.l) limit the AGV's velocity, acceleration, and angular velocity. (2.b) and (2.c) being nonlinear constraints, they are converted into soft constraints, resulting in a modified objective function:
\begin{equation}
\label{ex:min}
\setlength{\abovedisplayskip}{3pt}
\setlength{\belowdisplayskip}{3pt}
\begin{aligned}
&J=\left( T_f-T_0\right)+\varpi_1  \sum\nolimits_{{\text{i = 1}}}^{N-1} {{{\left( {{x_{i + 1}} - {x_i} - {v_i}\cdot \cos {{\theta _i}} \cdot {\Delta _{\text{t}}}} \right)}^2}} \\
&+\varpi_2 \sum\nolimits_{{\text{i = 1}}}^{N-1} {{{\left( {{y_{i + 1}} - {y_i} - {v_i} \cdot \sin  {{\theta _i}} \cdot {\Delta _{\text{t}}}} \right)}^2},}
\end{aligned}
\end{equation}
where $\varpi_1$ and $\varpi_2$ represent penalty function weights.
\section{ALGORITHM}
This section introduces obstacle handling, MVG for path planning, and the specific implementation of FSRC.
\subsection{Discretization Strategy for Obstacle Boundariesn}
\begin{figure}[htpb]
    \centering
    \subfloat[\label{fig:OBSTACLE1}]{
		\includegraphics[width=4cm]{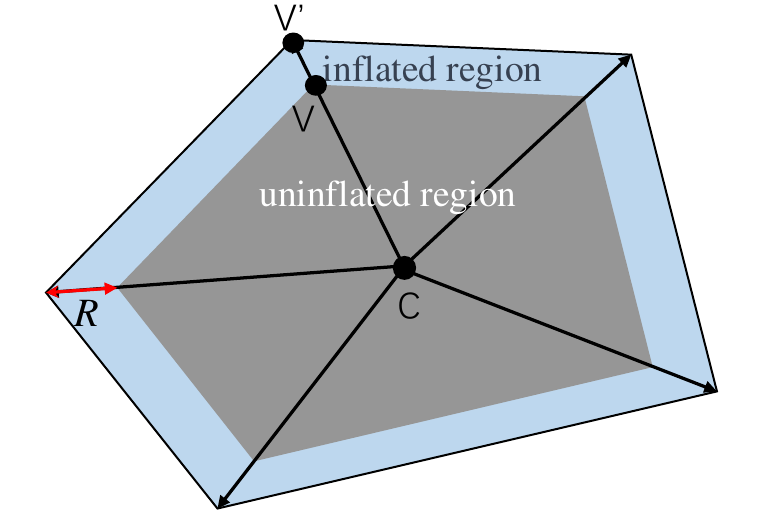}}
    \subfloat[\label{fig:OBSTACLE2}]{
		\includegraphics[width=4cm]{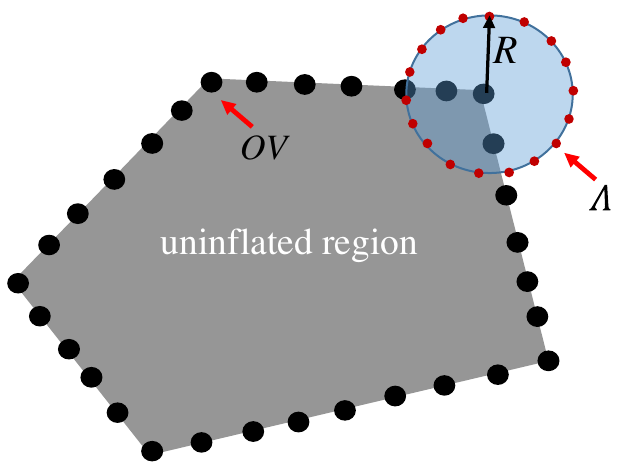}}
  \caption{Inflation of Obstacles and discretization}
  \label{fig:OBSTACLE}
\end{figure}

This section introduces an algorithm for discretizing obstacle boundaries, crucial for subsequent path planning and FSRC construction. The process involves gathering three sets: the expanded polygonal obstacle edges $\Xi$ and inflated obstacle vertex $\Psi$, utilized for obstacle avoidance of MVG, and the inflation obstacle node $\Lambda$, used to establish FSRC. These procedures are depicted in Algorithm~\ref{alg:oh}. Here, $R$ represents the coverage circle radius of the AGV (see Fig.~\ref{fig:scc}), $N_{obs}$ denotes the number of obstacles, $C_i$ represents the geometric center of obstacle $i$, and $(ox_{ij}, oy_{ij})$ denote the vertices of the obstacle. Fig.~\ref{fig:OBSTACLE} illustrates the process of obstacle inflation and discretization, Fig. \ref{fig:OBSTACLE1} demonstrates the acquisition of $\Xi$, and Fig. \ref{fig:OBSTACLE2} illustrates the acquisition of $\Lambda$. AGVs only need to avoid the boundaries of obstacles to ensure safety. Therefore, this paper focuses on discretizing only the boundaries of obstacles. Additionally, fewer inflation obstacle nodes lead to faster construction of FSRC.
\begin{spacing}{0.9}
\begin{algorithm}[htbp]
	\caption{Obstacle handling}
	\label{alg:oh}
	\begin{algorithmic}[1]
			\Require obstacles $O$, radius $R$, discretization precision $\ell$
            \State $\Xi \gets \emptyset $, $\Lambda  \gets \emptyset $, $OV  \gets \emptyset $, $\Psi \gets \emptyset$,
            \For {$i \gets 1$ to $N_{obs}$}
            \State $C_i \gets   {{\left( {\sum\nolimits_{j = 1}^{{N_{Vi}}} {o{x_{ij}}} ,\sum\nolimits_{j = 1}^{{N_{Vi}}} {o{y_{ij}}} } \right)} \mathord{\left/
 {\vphantom {{\left( {\sum\nolimits_{j = 1}^{{N_{Vi}}} {o{x_{ij}}} ,\sum\nolimits_{j = 1}^{{N_{Vi}}} {o{y_{ij}}} } \right)} {{N_{Vi}}}}} \right.
 \kern-\nulldelimiterspace} {{N_{Vi}}}}$
            \For {$j \gets 1$ to $N_{Vi}$}
            \State ${V_{ij}^{'}} \gets $ expand $\left( R,V_{ij} \right)$ in the direction $\overrightarrow {C_{i}V_{ij}}$
            \State add  ${V_{ij}^{'}} \gets $ to $\Psi$
            \EndFor
            \State add  $[{V_{i,0}^{'}},{V_{i,N_{Vi}}^{'}}]$ to $\Xi$
            \For  {$j \gets 1$ to $N_{Vi}-1$}
            \State add  $[{V_{i,j}^{'}},{V_{i,j+1}^{'}}]$ to $\Xi$
            \EndFor
            \State $OV \gets $ {discretizeBoundary} $\left(O_i, \ell\right)$
            \State $\Lambda_i \gets $ {inflateBoundary} $\left(OV, R ,\ell\right)$
            \State add  $\Lambda_i$ to $\Lambda$
            \EndFor
			\State \Return{$\Lambda, \Xi \text{ and } \Psi$}
	\end{algorithmic}
\end{algorithm}
\end{spacing}
\subsection{Modified Visibility Graph}
\begin{figure}[htbp]
      \centering
      \setlength{\abovecaptionskip}{0.cm}
      \includegraphics[width=7.4cm]{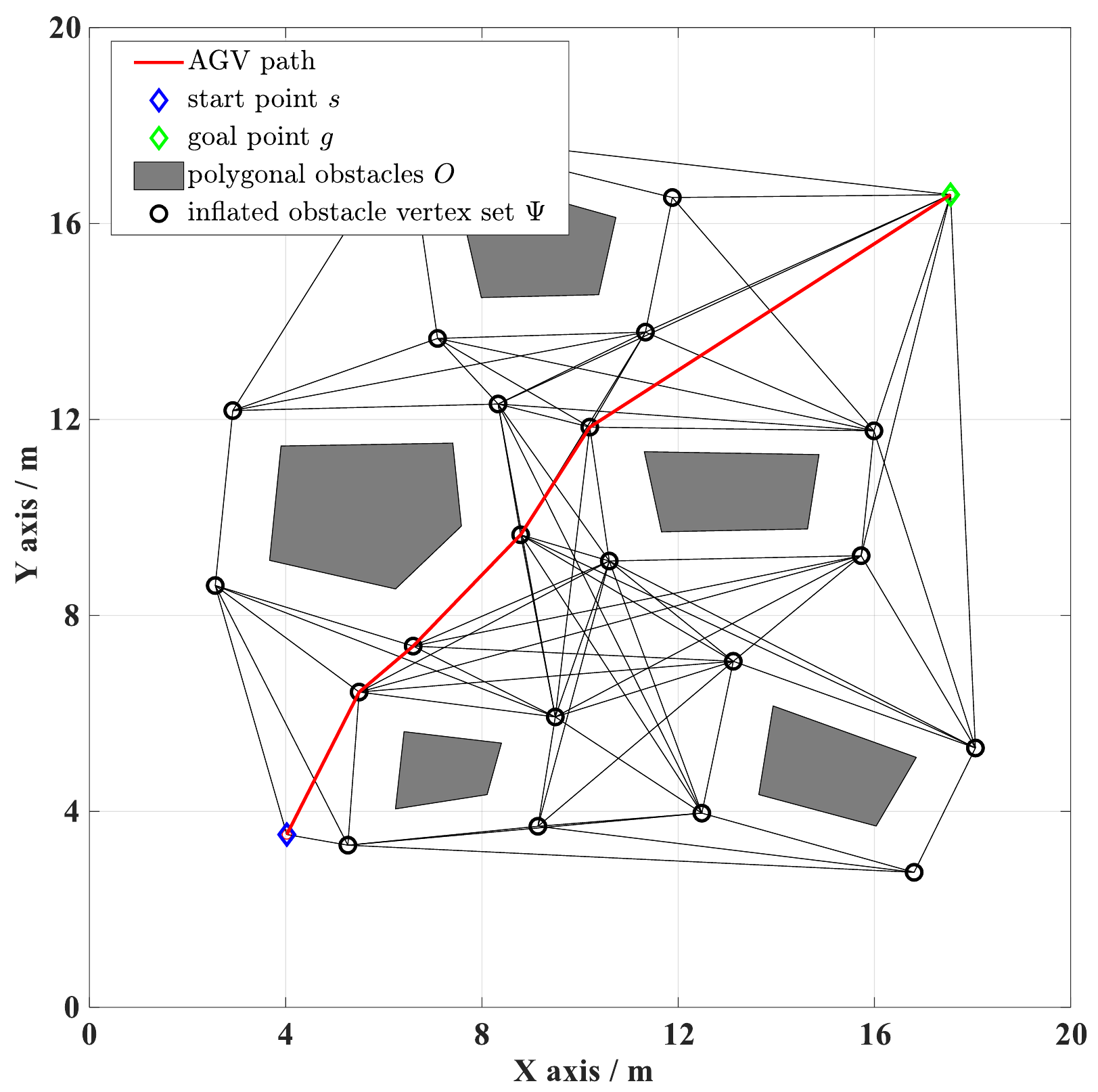}
      \caption{
      Results of the Modified Visibility Graph. (Enhance visibility for finer details within the figure by zooming in.)}
      \label{AGVPATH}
\end{figure}

Building upon \cite{lozano1979algorithm}, the Modified Visibility Graph (MVG) algorithm for path planning is introduced, as depicted in Algorithm~\ref{alg:MVG}. This approach encompasses three primary steps. Firstly, a bidirectional weighted graph G is created using the start point $s$, the goal point $g$, the expanded polygonal obstacle edges $\Xi$, and the inflated obstacle vertex set $\Psi$. A compact adjacency list $\mathfrak{L}$ is employed to efficiently represent this graph G. The weights between nodes are distance-based, and infinite values indicate collisions. Collision-free paths are delineated by line segments that do not intersect any line segment within $\Xi$. Secondly, the Dijkstra search \cite{dijkstra2022note} focuses on finding the shortest path between $s$ and $g$ rather than examining all possible node pairs. The search concludes once the shortest path between $s$ and $g$ is identified. Thirdly, the outcomes of the second phase are uniformly discretized into $N$ path points. The example can be referenced in Fig.~\ref{AGVPATH}.
\begin{spacing}{0.9}
\begin{algorithm}[htbp]
	\caption{Modified Visibility Graph (MVG)}\label{alg:MVG}
	\begin{algorithmic}[1]
			\Require start point $s$, goal point $g$, $\Xi$, $\Psi$, number of path points $N$
            \State $ \mathfrak{L}\gets $ {constructAdjacencyList} $\left(s, g, \Xi, \Psi \right)$
            \State $ \left(pX, pY\right) \gets $ {Dijkstra} $\left(\mathfrak{L}, s, g\right)$
            \State $ path \gets $ {discretePath} $\left(pX, pY, N\right)$
			\State \Return{$path$}
	\end{algorithmic}
\end{algorithm}
\end{spacing}
\subsection{Fast Safe Rectangular Corridor}
\begin{spacing}{0.9}
	\begin{algorithm}[htbp]
		\caption{Fast Safe Rectangular Corridor (FSRC)}
		\label{alg:SRC}
		\begin{algorithmic}[1]
			\Require path points $path(X,Y)$, inflation obstacle node set $\Lambda$, number of discrete path points $N$
			\State Initialize parameter $\tau, \gamma$, $L_m$, $\chi$, $\mathcal{B_R}$ and $T_m$
			\For {$i \gets 1$ to $N$}
			\State $k \gets 1, x_i \gets X(i), y_i \gets Y(i)$
			\If {$i>1$ and $k \leq T_m \ \AND  \ (x_i,y_i) \ \text{in} \ \mathcal{B_R}(i-1,:)$}
			\State  $\mathcal{B_R}(i,:) \gets \mathcal{B_R}(i-1,:), k \gets k+1$
			\State  \textbf{continue}
			\EndIf
			\State $k \gets 1 $
			\State $\mathcal{B_R}(i,:) \gets [L_{m},L_{m},L_{m},L_{m}] $
			\State $\Lambda_{rm}\gets $ {$\texttt{obsCheck}(\mathcal{B_R}(i,:), \Lambda,x_i,y_i)$}
			\If {$\Lambda_{rm}=\emptyset$}
			\State  \textbf{continue}
			\EndIf
			\State $\mho \gets [0,0,0,0]$, $box \gets [\tau,\tau,\tau,\tau] $
			\State $\mathcal{B_R}(i,:) \gets [0,0,0,0] $
			\While {sum$(\mho) <4 $ }
			\For {$j \gets 1$ to $4$}
			\If {$\mho(j)=1$}
			\State  \textbf{continue}
			\EndIf
			\State $tmp \gets \mathcal{B_R}(i,:), tmp(j) \gets tmp(j)+box_j  $
			\If {$tmp(j) > L_m$}
			\State  $\mho(j) \gets 1$, \textbf{continue}
			\EndIf
			\State $\tilde{\Lambda_{rm}}\gets $ {$\texttt{obsCheck}(\mathcal{B_R}(i,:), \Lambda_{rm},x_i,y_i)$}
			\If {$\tilde{\Lambda_{rm}}=\emptyset$}
			\State  $\mathcal{B_R}(i,j) \gets tmp(j)$, $box_j \gets box_j \times \gamma$
			\Else
			\State $box_j \gets \min(box_j/4,\chi)$
			\State $tmp(j) \gets \mathcal{B_R}(i,j)+box_j$
			\While{$tmp(j)<L_m$}
			\State $\tilde{\Lambda_{rm}}\gets $ {$\texttt{obsCheck}(\mathcal{B_R}(i,:), {\Lambda_{rm}})$}
			\If {$\tilde{\Lambda_{rm}}=\emptyset$}
			\State $\mathcal{B_R}(i,j) \gets tmp(j)$
			\State $tmp(j) \gets tmp(j)+box_j$
			\Else
			\State  break
			\EndIf
			\EndWhile
			\State  $\mho(j) \gets 1$
			\EndIf
			\EndFor
			\EndWhile
			\EndFor
			\State \Return{$\mathcal{B_R}$}
		\end{algorithmic}
	\end{algorithm}
\end{spacing}
In this section, we introduce the establishment process of FSRC, which consists of $N$ rectangular boxes $\mathcal{B_R}(i,:)$, where $i = 1,2,...,N$. Each rectangular box $\mathcal{B_R}(i,:)$ is generated by extending from the path point $P_i$, and it must not contain any obstacle nodes from $\Lambda$. $\mathcal{B_R}(i,:)$ represents the lengths of the extension from ${P_i}$ in four directions. The $\left( {x_i^{\max },x_i^{\max }} \right)$ and $\left( {y_i^{\min },y_i^{\max }} \right)$ for \texttt{Problem} (\ref{ex:OM}) can be calculated from the rectangular box $\mathcal{B_R}(i,:)$ and ${P_i}$
\begin{equation}
	\label{ex:rb}
	\setlength{\abovedisplayskip}{3pt}
	\setlength{\belowdisplayskip}{3pt}
	\begin{aligned}
	 &{B_R}\left( {i,:} \right) = ({L_1},{L_2},{L_3},{L_4}) \hfill \\
	&x_i^{\min } = {x_i} - {L_2},x_i^{\max } = {x_i} + {L_4} \hfill \\
	&y_i^{\min } = {y_i} - {L_1},y_i^{\max } = {y_i} + {L_3} \hfill \\
	\end{aligned}
\end{equation}

\begin{figure}[htbp]
	\centering
	\subfloat[\label{fig:SRC1}]{
		\includegraphics[width=4cm]{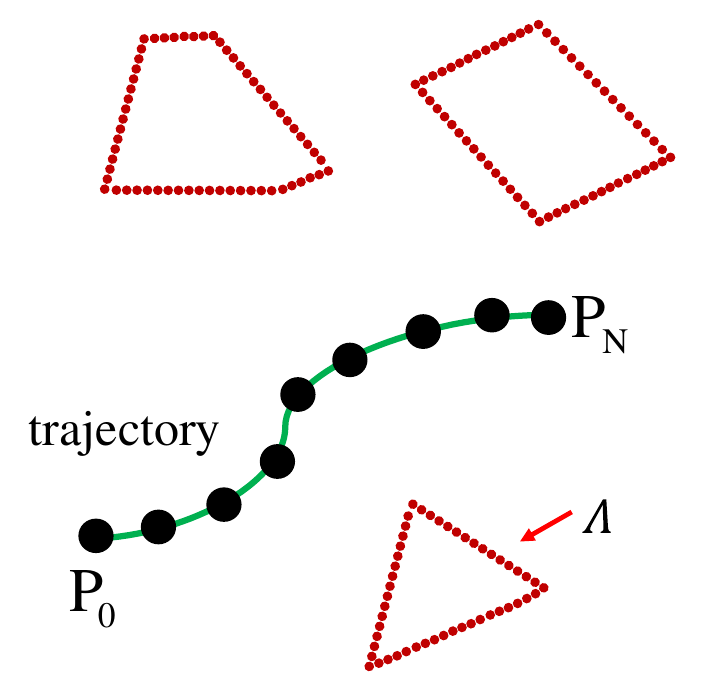}}
	\subfloat[\label{fig:SRC2}]{
		\includegraphics[width=3.5cm]{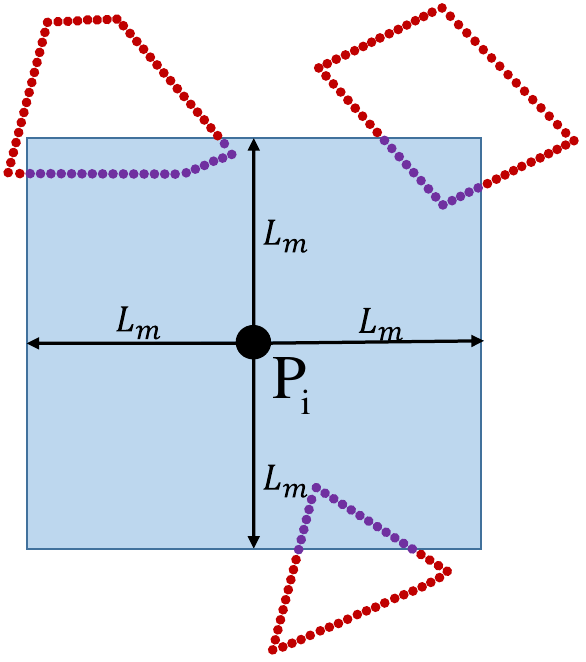}}\\
	\subfloat[\label{fig:SRC3}]{
		\includegraphics[width=4cm]{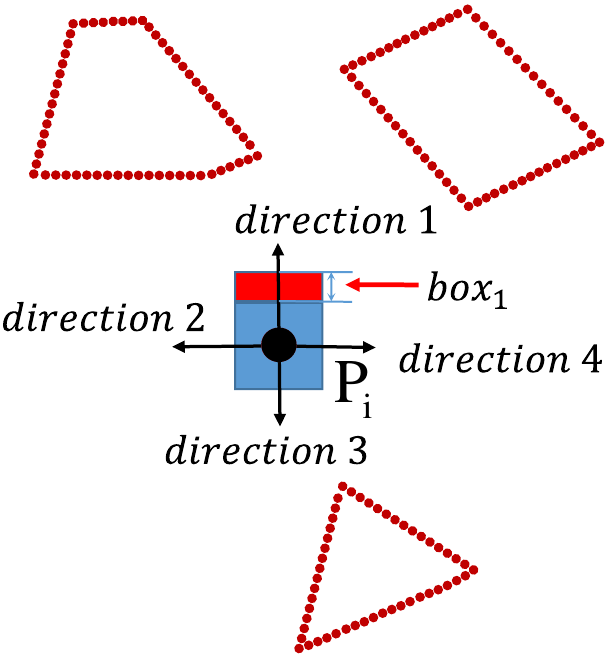}}
	\subfloat[\label{fig:SRC4}]{
		\includegraphics[width=4cm]{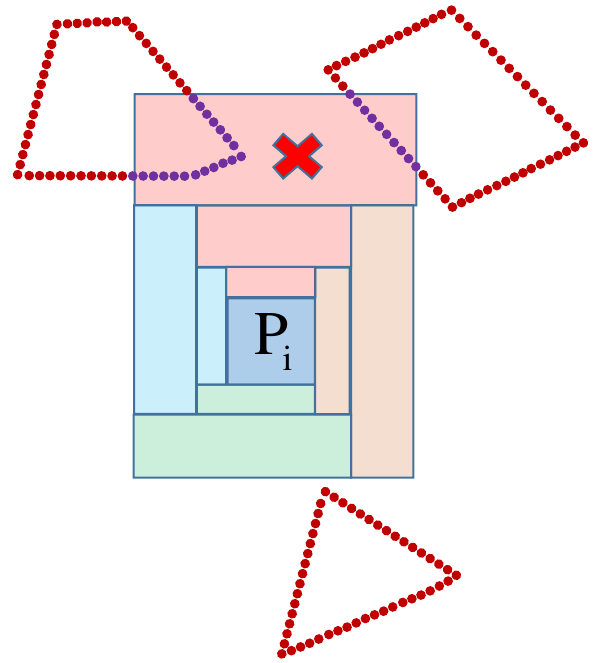}}
	\caption{The creation process of FSRC:
		(a) path points and obstacle set $\Lambda$,
		(b) initial bounding box $\mathcal{B_R}(i,:)$,
		(c) expansion in four directions,
		(d) expansion encountering obstacles.}\label{fig:SRC}
\end{figure}

Figure~\ref{fig:SRC} depicts the construction process of the rectangular corridor, utilizing the discrete path point sequence $path$ generated by Algorithm~\ref{alg:MVG} and $\Lambda$ obtained from Algorithm~\ref{alg:oh}. Figures~\ref{fig:SRC} (b)-(d) illustrate the process of generating the corresponding rectangular box $\mathcal{B_R}(i,:)$ for ${P_i}({x_i},{y_i})$.
Starting from ${P_i}({x_i},{y_i})$, the box is iteratively expanded in the up, left, down, and right directions (denoted as directions 1-4), with each expansion adding an extension rectangle to the box in that direction. Algorithm~\ref{alg:SRC} outlines the process of generating FSRC, which is expedited by the following three strategies.

\textbf{Strategy 1}: The bounding boxes of two adjacent path points $P_i$ and $P_{i+1}$ often exhibit overlapping tendencies owing to their proximity. A check is conducted to avoid redundant box creation: if $P_i$ is within the boundaries of the previous bounding box, $\mathcal{B_R}(i-1,:)$, the current process is skipped. This omission step can be repeated up to a maximum of $T_m$ times.

\textbf{Strategy 2}: Initialize a rectangular box denoted as $\left[ {{L_m},{L_m},{L_m},{L_m}} \right]$, where ${L_m}$ represents the maximum allowable extension length. If no obstacle nodes are within the initial box, subsequent operations are skipped to reduce the construction time. However, if obstacles are detected inside the initial box, expansion starts from point ${P_i}({x_i},{y_i})$, ensuring that the subsequent expansion does not exceed the range of the initial box, i.e., ${L_i} \leqslant {L_m}$ for $i = 1,2,3,4$. Moreover, use the obstacle nodes inside the initial box, denoted as ${\Lambda _{rm}} \subset \Lambda$, for obstacle detection using $\texttt{obsCheck}()$, instead of using all nodes in $\Lambda$. Limiting ${L_i}$ to expand at most up to the boundary ${L_m}$ accelerates the execution speed of the $\texttt{obsCheck}()$ function.
\begin{figure}[htbp]
	\centering
	\subfloat[\label{fig:IS}]{
		\includegraphics[width=4cm]{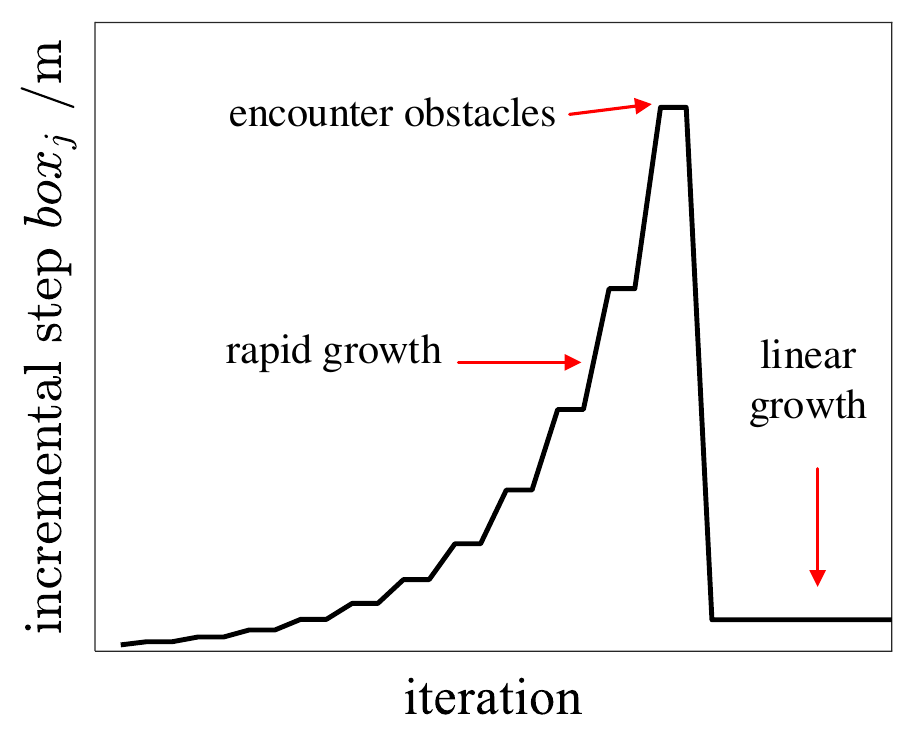}}
	\subfloat[\label{fig:EL}]{
		\includegraphics[width=4cm]{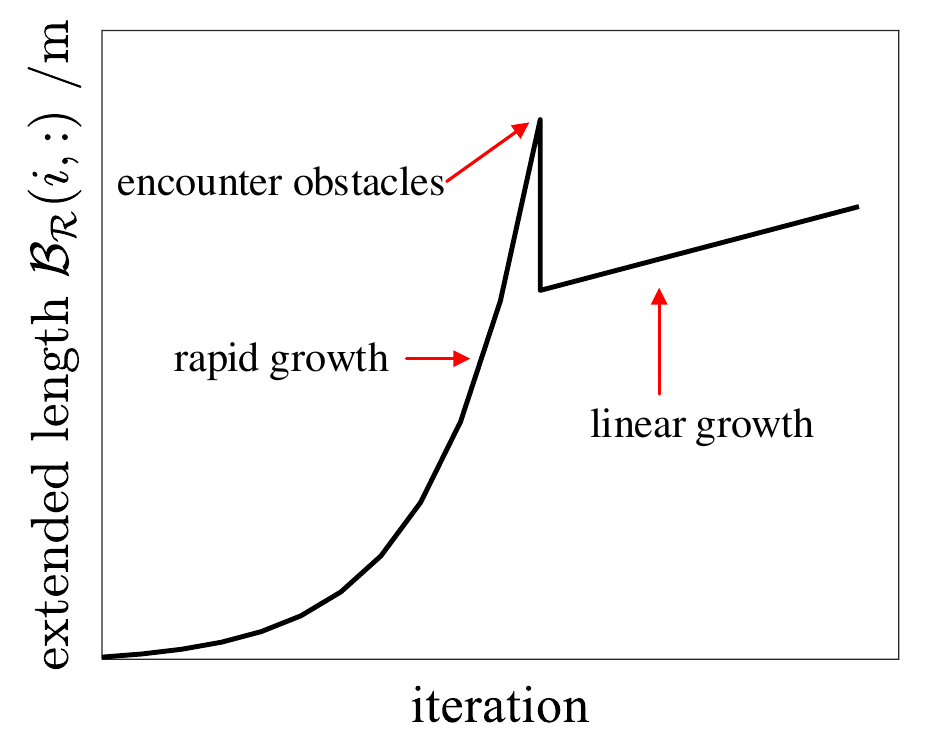}}
	\caption{Rapid growth and linear growth.}\label{fig:SrcExpand}
\end{figure}

\textbf{Strategy 3}: We don't use uniform expansion to quickly reach the obstacle boundary. Instead, we divide expansion into two stages for each direction: rapid growth and linear growth, using a step size of $box$ with an initial value of $\tau$.
\begin{itemize}
	\item \textbf{Rapid Growth}: The expansion length $box_j$ in each direction is doubled by a factor of $\eta$ in each iteration until it reaches the obstacle boundary or the length of ${L_m}$. If the expansion encounters an obstacle or exceeds $L_m$, the step size $box_j$ is adjusted to the minimum value between $box_j/4$ and $\chi$.
	\item \textbf{Linear Growth}: Expansion in the current direction continues with a constant step size $box_j$ until a collision with an obstacle or reaching $L_m$ occurs.
\end{itemize}
The relationship between the expansion length $box_j$ and the accumulated expansion length $\mathcal{B_R}(i-1,:)$ with iteration number is illustrated in Fig.~\ref{fig:SrcExpand}.

By employing these three strategies, we reduce the FSRC generation time, enhancing the overall efficiency of the trajectory planning framework.
\section{EXPERIMENT RESULTS AND DISCUSSIONS}
\begin{table}[htbp]
	\footnotesize
	\centering
    \setlength{\tabcolsep}{0pt}
	\caption{\label{tab:ParametersTab} Parameters of the experiment.}
    \begin{center}
	\begin{tabular}{lclll}
		\toprule
		Parameter   & Value \\
		\midrule
		AGV Length  &0.612 m	\\
		\midrule
		 AGV Width &0.582 m\\
		\midrule
		 maximum speed  ${v}^{\max}$ & 3.0 m$/s$  \\
		\midrule
		maximum acceleration  ${a}^{\max}$  & 1.8 m$/s^2$  \\
		\midrule
		maximum angular velocity  ${\omega}^{\max}$  & 2.5 $rad/s$  \\
		\midrule
		map size  & 20 $\times$ 20 m  \\
		\midrule
		number of discrete points $N$ for \texttt{Problem} (\ref{ex:OM})  & 80  \\
		\midrule
		discretization precision of obstacle points $\ell$  & 0.1 m  \\
		\midrule
		half-length of initial box $L_{m}$  & 10 m  \\
		\midrule
		maximum repetition count $T_m$   & 8  \\
		\midrule
		growth factor $\gamma$  & 2  \\
		\midrule
		minimum growth length $\chi$  & 0.2 m  \\
		\midrule
		grid map precision ($x, y, \theta$) for hybrid A*  & 0.1 m, 0.1 m, 0.1 $rad$  \\
		\bottomrule
	\end{tabular}
    \end{center}
\end{table}

\begin{figure}[htbp]
	\centering
	\includegraphics[width=8.4cm]{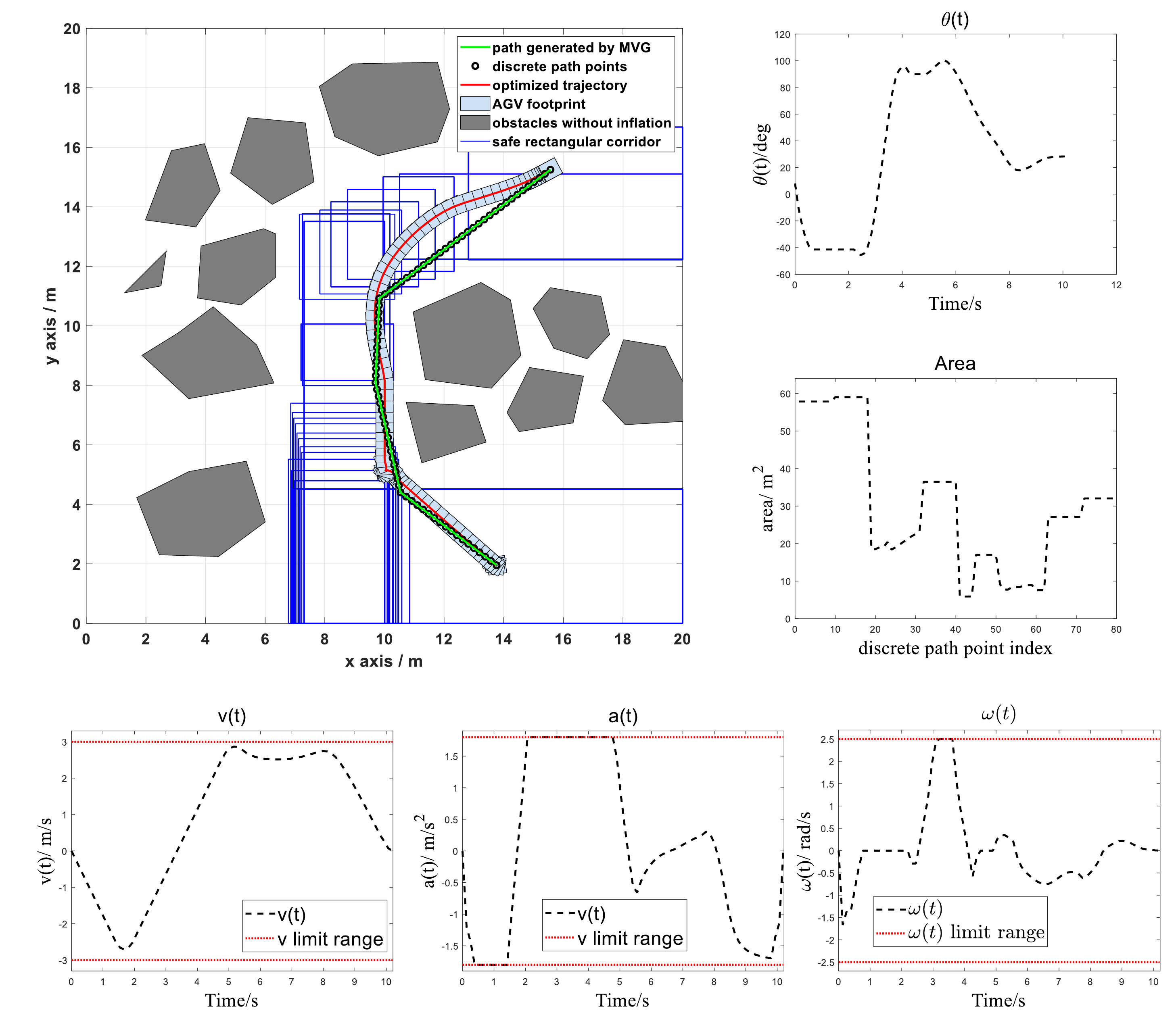}
	\caption{
		Simulation results for trajectory planning in Case1.}
	\label{fig:case1}
\end{figure}

\begin{figure*}[htbp]
	\centering
	\subfloat[\label{fig:case2}]{
		\includegraphics[width=4cm]{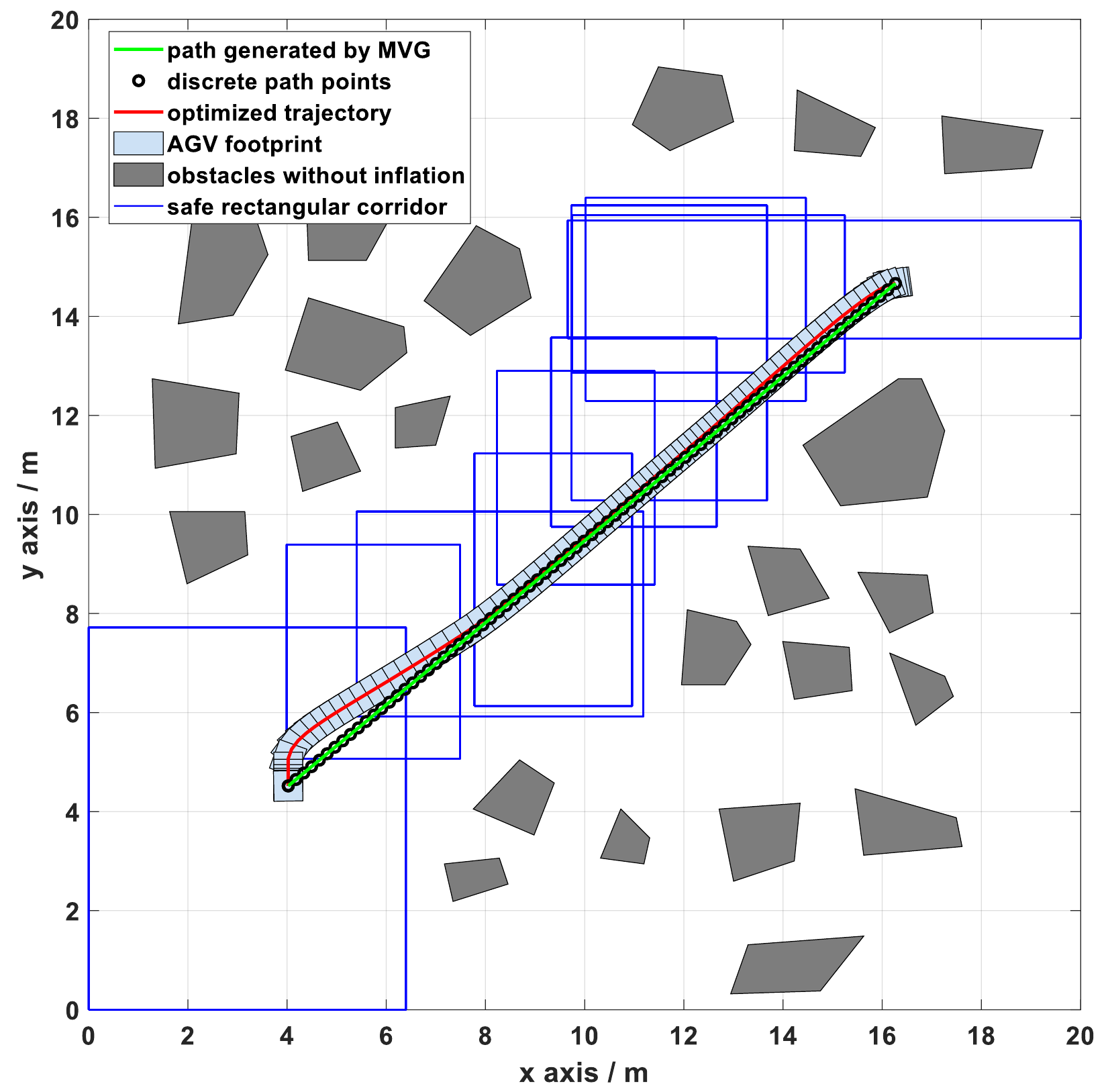}}
	\subfloat[\label{fig:case3}]{
		\includegraphics[width=4cm]{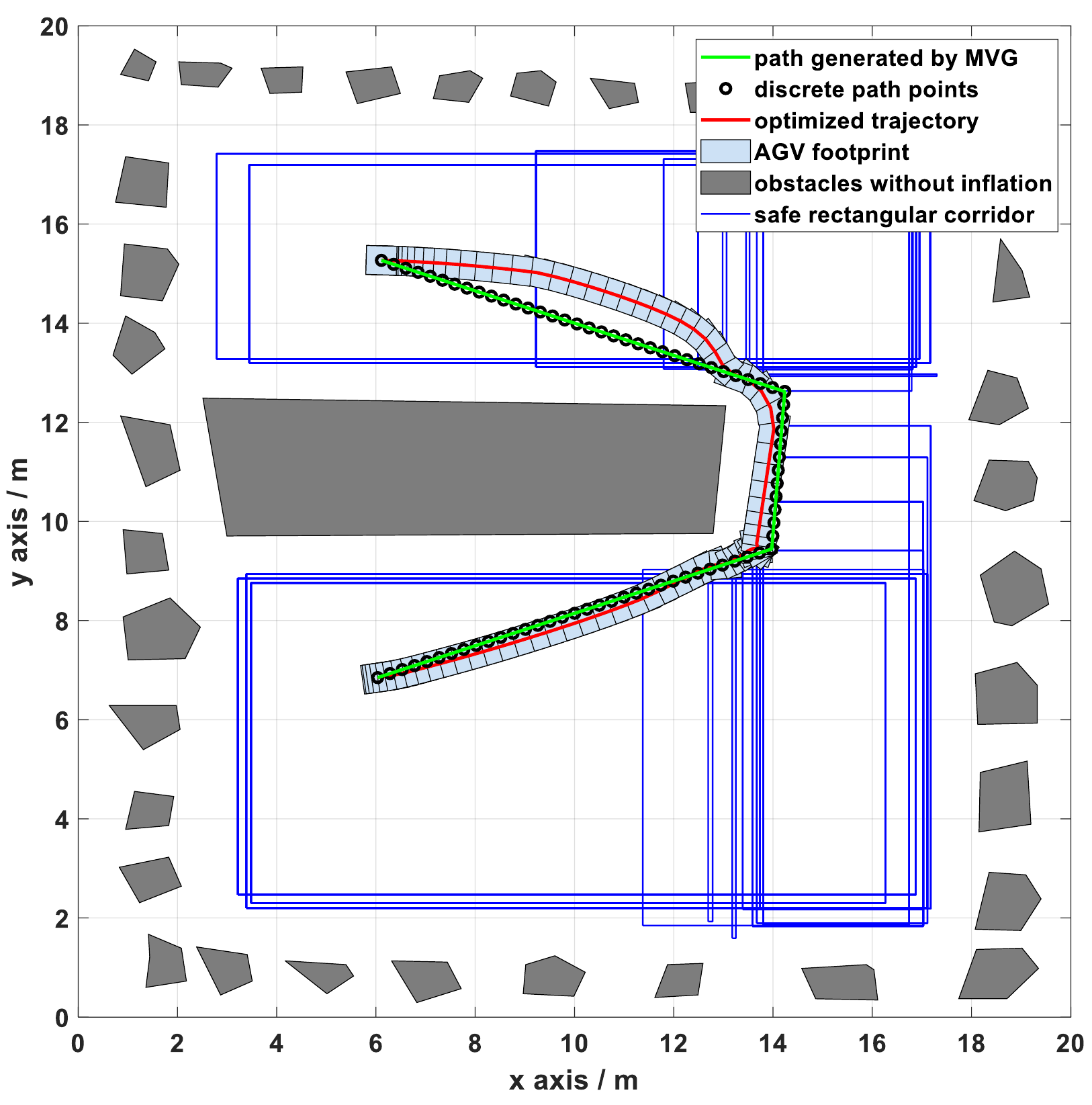}}
	\subfloat[\label{fig:case2}]{
		\includegraphics[width=4cm]{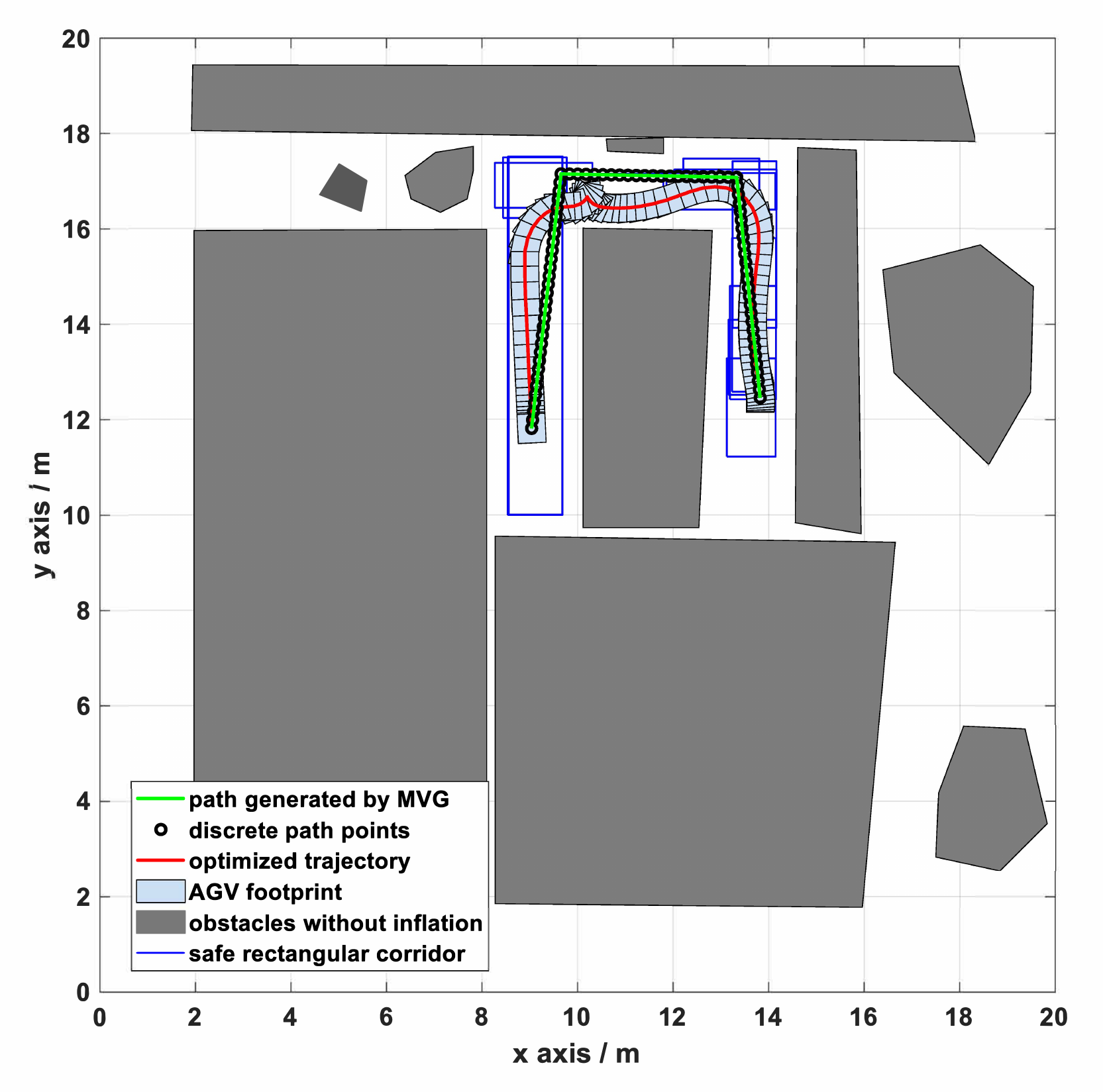}}
	\subfloat[\label{fig:case3}]{
		\includegraphics[width=4cm]{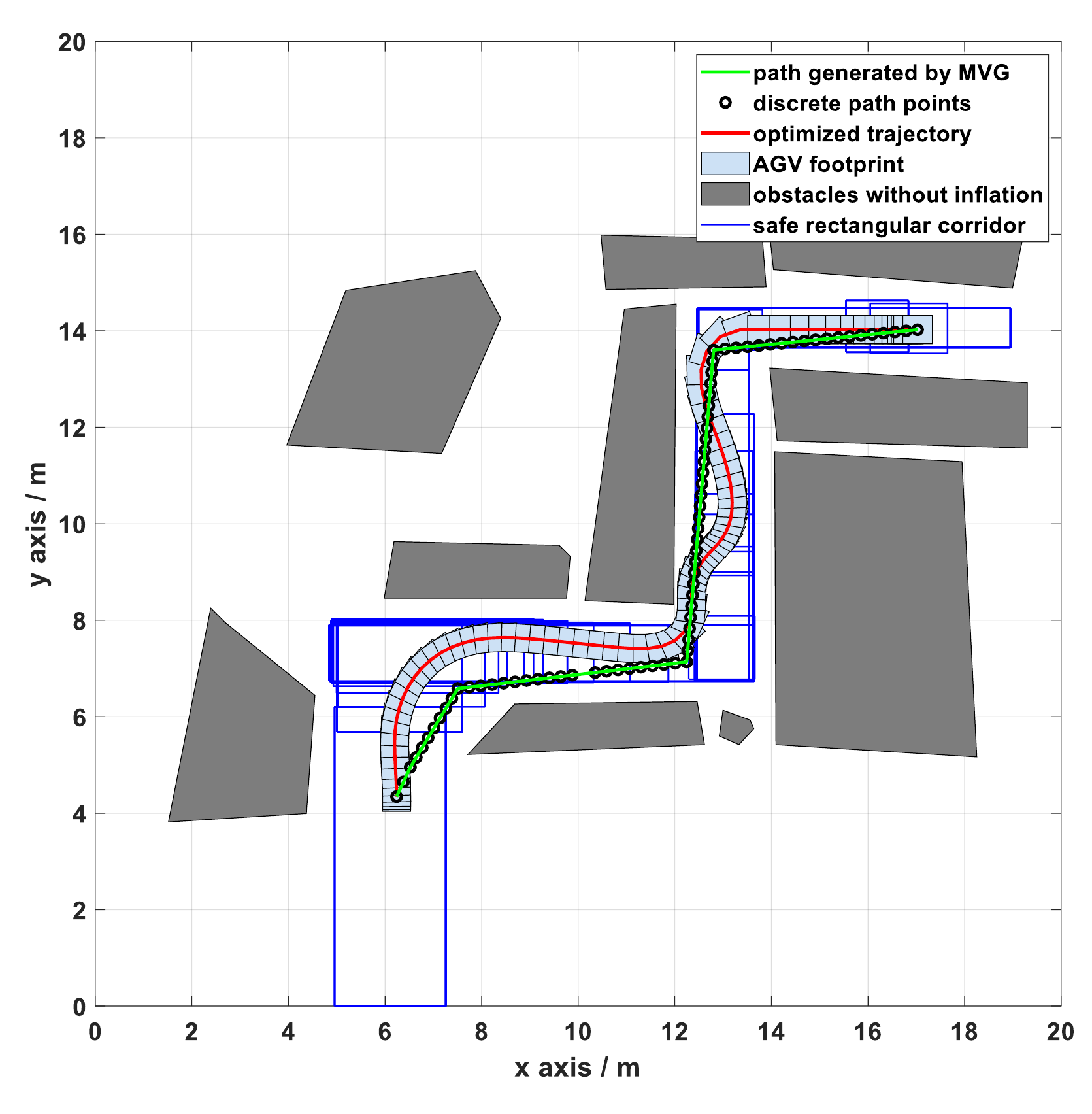}}
	\caption{
		Simulation results for trajectory planning: (a) Case2, (b) Case3, (c) Case4, (d) Case5.}\label{fig:case23}
\end{figure*}

This section presents simulation and physical experiments. We demonstrate the computational efficiency of our FSRC algorithm compared to other safe convex corridor-based methods. Additionally, we compare our framework with advanced trajectory planning frameworks in simulation experiments to validate its exceptional performance. Furthermore, we validate the practical viability of our framework through physical platform experiments. Detailed parameter configurations for experiments are provided in Table \ref{tab:ParametersTab}.
\subsection{Simulation}
In our simulation experiments, we utilized IPOPT \cite{wachter2006implementation} and MA57 \cite{hsl2007collection} as the solvers for the \texttt{Problem} (\ref{ex:OM}). These experiments were carried out using MATLAB 2020b on a desktop computer with an AMD Ryzen 5 3600X 6 Core CPU running at 3.8 GHz and 16 GB of RAM.

The simulation results of our proposed framework have been presented in five distinct scenarios, as illustrated in Fig.\ref{fig:case1} and Fig.\ref{fig:case23}, which can be zoomed in to observe details. These scenarios encompass varying obstacle densities: Case1 (low), Case2 (medium), and Case3 (high), along with Case4 and Case5 depicting narrow passages. In all five cases, the blue rectangular boxes generated by FSRC ensure that the AGV avoids contact with obstacles. Specifically, Fig.~\ref{fig:case1} illustrates that the AGV's speed, acceleration, and angular velocity remained within predefined ranges. Additionally, the area figure demonstrates the area of each rectangular box $\mathcal{B_R}_i$ generated by FSRC.
\begin{table}[htbp]
	\centering
	\footnotesize
	\setlength{\tabcolsep}{3pt}
	\caption{Efficiency comparison of Safe Convex Corridor-based methods (Time: seconds).}
	\begin{tabular}{cccccc}
		\toprule
		Method & Scenario 1 & Scenario 2  & Scenario  3 & Scenario 4  & Mean \\
		\midrule
		FSRC  & \textbf{0.0971}&\textbf{0.0959}&\textbf{0.0825}& \textbf{0.0855}& \textbf{0.0903}\\
		STC   & 0.5766&1.3990&0.5034&0.5014&0.7451	 \\
		SFC   & 0.2193&0.2288&0.2910&0.2109& 0.2375 \\
		\bottomrule
	\end{tabular}
	\label{tab:FRSC}
\end{table}

To validate the computational efficiency of FSRC, we compared it with two other advanced methods based on safe convex corridors, STC \cite{li2020maneuver} and SFC \cite{li2020maneuver}. FSRC outperformed both, with average efficiency improvements of 725\% and 163\%, respectively, as shown in Table \ref{tab:FRSC}. This superiority is attributed to the three strategies implemented to minimize corridor construction time. SFC constructs convex polygons or polytopes, and its process is complex, leading to complexity and lower efficiency. STC uses uniform expansion, with equal lengths for each expansion, which is less efficient than our two-stage approach: rapid and linear growth. Additionally, the FSRC algorithm proposed by us utilizes obstacle boundary points for collision detection in corridor construction, instead of the entire obstacle area, further enhancing the construction speed.
\begin{table}[htbp]
	\centering
	\footnotesize
	\setlength{\tabcolsep}{3pt}
	\caption{ Efficiency comparison of trajectory planning frameworks (Time: seconds).}
	\begin{center}
		\begin{tabular}{c|ccccccc}
			\toprule
			\multicolumn{1}{c}{case} & \textbf{Method}    &\textbf{Mapping}  &\textbf{Path} & \textbf{Corridor} & \textbf{Optimization}& \multicolumn{1}{c}{\textbf{Total}} \\
			\midrule
			\multirow{3}[1]{*}{Case1} & Proposed &  \textbf{0.143} &  \textbf{0.004} &  \textbf{0.103} &  \textbf{0.416} &  \textbf{0.667} \\
			& Area-based   & 0.387 & 11.531 & -     & 21.560 & 33.477 \\
			& STC   & 0.387 & 11.531 & 0.473 & 1.632 & 14.021 \\
			\midrule
			\multirow{3}[1]{*}{Case2} & Proposed &  \textbf{0.285} &  \textbf{0.004} &  \textbf{0.075} &  \textbf{0.375} &  \textbf{0.739} \\
			& Area-based   & 0.303 & 0.276 & -     & 48.055 & 48.634 \\
			& STC   & 0.303 & 0.276& 0.440  & 0.811 & 1.830 \\
			\midrule
			\multirow{3}[1]{*}{Case3} & Proposed &  \textbf{0.551} &  \textbf{0.004} &  \textbf{0.112} &  \textbf{0.380} &  \textbf{1.049} \\
			& Area-based   & 0.354 & 28.093 & -     & 5.571 & 34.018 \\
			& STC   & 0.354 & 28.093 & 0.651 & 0.861 & 29.605 \\
			\midrule
			\multirow{3}[1]{*}{Case4} & Proposed &  \textbf{0.08} &  \textbf{0.004} &  \textbf{0.112} &  \textbf{0.364} &  \textbf{0.596} \\
			& Area-based   &  0.639 & 0.508 & -     & 3.88 & 5.027 \\
			& STC   & 0.639 & 0.508 & 0.279 & 0.383 & 1.809 \\
			\midrule
			\multirow{3}[1]{*}{Case5} & Proposed &  \textbf{ 0.093} &  \textbf{0.004} &  \textbf{0.194} &  \textbf{0.378} &  \textbf{ 0.669} \\
			& Area-based   &  0.344 & 1.891 & -     & 4.95 & 7.185 \\
			& STC   &  0.344 & 1.891 & 0.301 & 0.363 & 5.537 \\
			\bottomrule
		\end{tabular}
		\label{tab:res}
	\end{center}
\end{table}

\begin{table}[htbp]
	\centering
	\footnotesize
	\setlength{\tabcolsep}{3pt}
	\caption{The comparison of the optimal path lengths.}
	\begin{tabular}{ccccccc}
		\toprule
		Method & Case1 & Case2  & Case3 & Case4  & Case5 & Mean \\
		\midrule
		Proposed    & 18.952 & 21.340 & 21.185 & 13.404 & 19.047 & 18.786 \\
		Area-based  & 18.724 & 34.919 & 19.254 & 12.247 & 18.964 & 20.822 	 \\
		STC         & 19.456 & 15.906 & 20.397 & 14.545 & 17.971 & 17.655  \\
		\bottomrule
	\end{tabular}
	\label{tab:LEN}
\end{table}

Our proposed framework was comprehensively evaluated by comparing it with two advanced frameworks: the area-based method \cite{li2015unified} and the STC-based method \cite{li2020maneuver}. Table \ref{tab:res} summarizes the computational times for these three frameworks. In this table, ``Mapping" represents the map construction time. Our framework constructs an adjacency list (Algorithm~\ref{alg:oh}), while the others make grid maps. ``Path" represents the path planning time. We use MVG, while the others use hybrid A*, thus requiring grid map construction. ``Corridor" represents the time for safe corridor construction. Since the area-based method does not construct corridors, its time is 0. Significantly, our proposed algorithm outperforms the others in multiple aspects, including path planning, optimization problem-solving, and total computation time, highlighting the superior performance of our framework. This highlights the superior performance of our framework. Compared to these two frameworks, our framework achieves computational efficiency gains of 1 to 2 orders of magnitude.

Finally, Table \ref{tab:LEN} shows the optimal path lengths obtained by the three frameworks in the five cases. The differences in the shortest paths obtained by the three frameworks are insignificant, with STC having the shortest path.
\subsection{PHYSICAL EXPERIMENT}
\begin{figure}[htbp]
   \centering
      \setlength{\abovecaptionskip}{0.cm}
      \includegraphics[width=6cm]{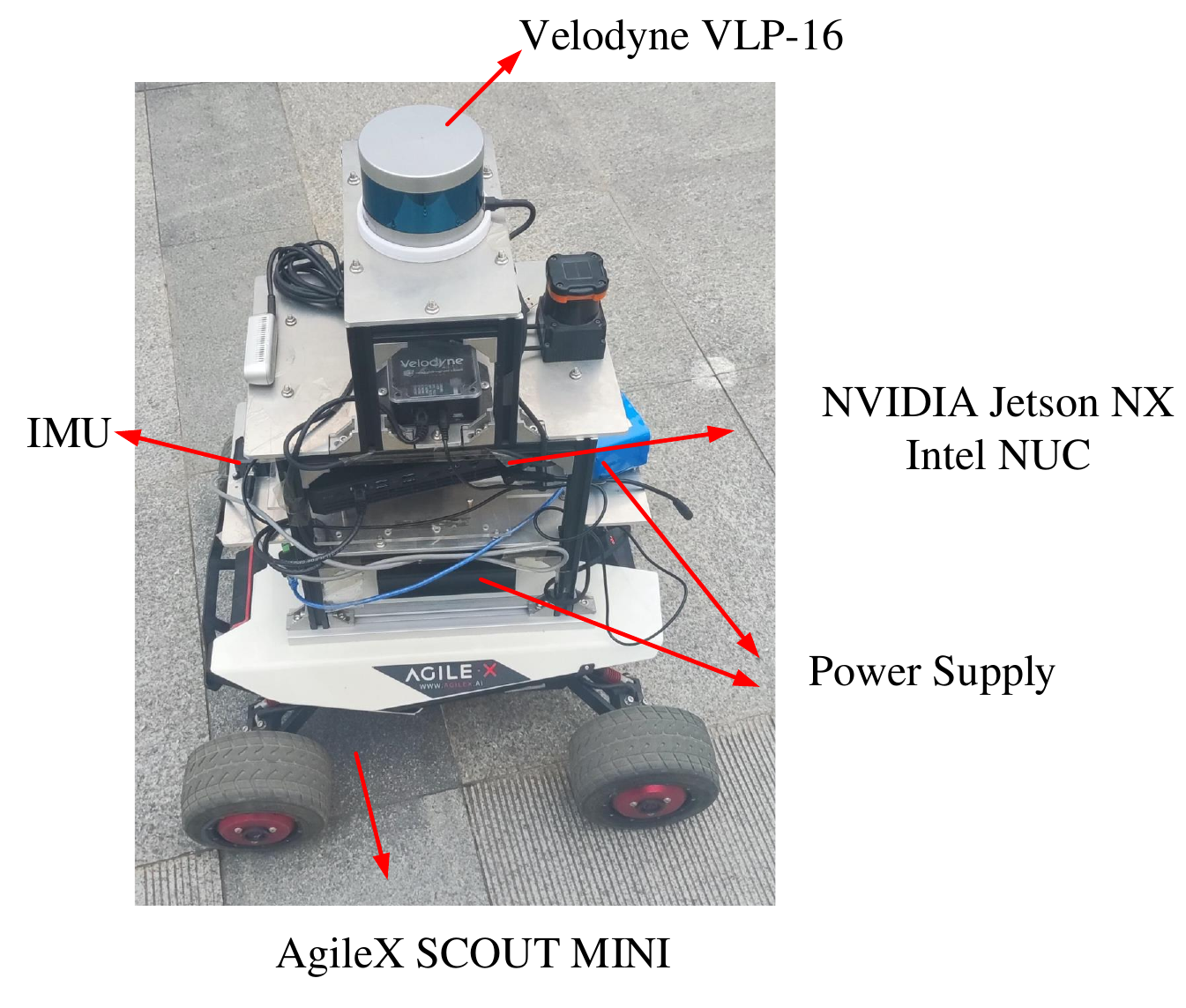}
      \caption{
      Physical experiment environment with AgileX SCOUT MINI platform.}
      \label{fig:NEWPE}
\end{figure}

Our physical experiment AGV platform is built upon the AgileX SCOUT MINI robot chassis development platform (refer to Fig.~\ref{fig:NEWPE}). This platform offers exceptional terrain adaptability and ground clearance, enabling agile movement across various surfaces. The fundamental parameters of the SCOUT MINI can be found in Table \ref{tab:ParametersTab}. The AGV platform is equipped with a Velodyne VLP-16 3D LiDAR with a maximum range of 100m, a 9-axis IMU, and an NVIDIA Jetson NX. The NVIDIA Jetson NX features a GPU with 384 cores (Volta architecture @1100MHz + 48 Tensor Cores) and a CPU with NVIDIA Carmel ARMv8.2 (6-core) @1.4GHz (6MB L2 + 4MB L3 cache).

\begin{figure}[htbp]
   \centering
   \subfloat[\label{fig:Actual_Map}]{
		\includegraphics[width=4cm]{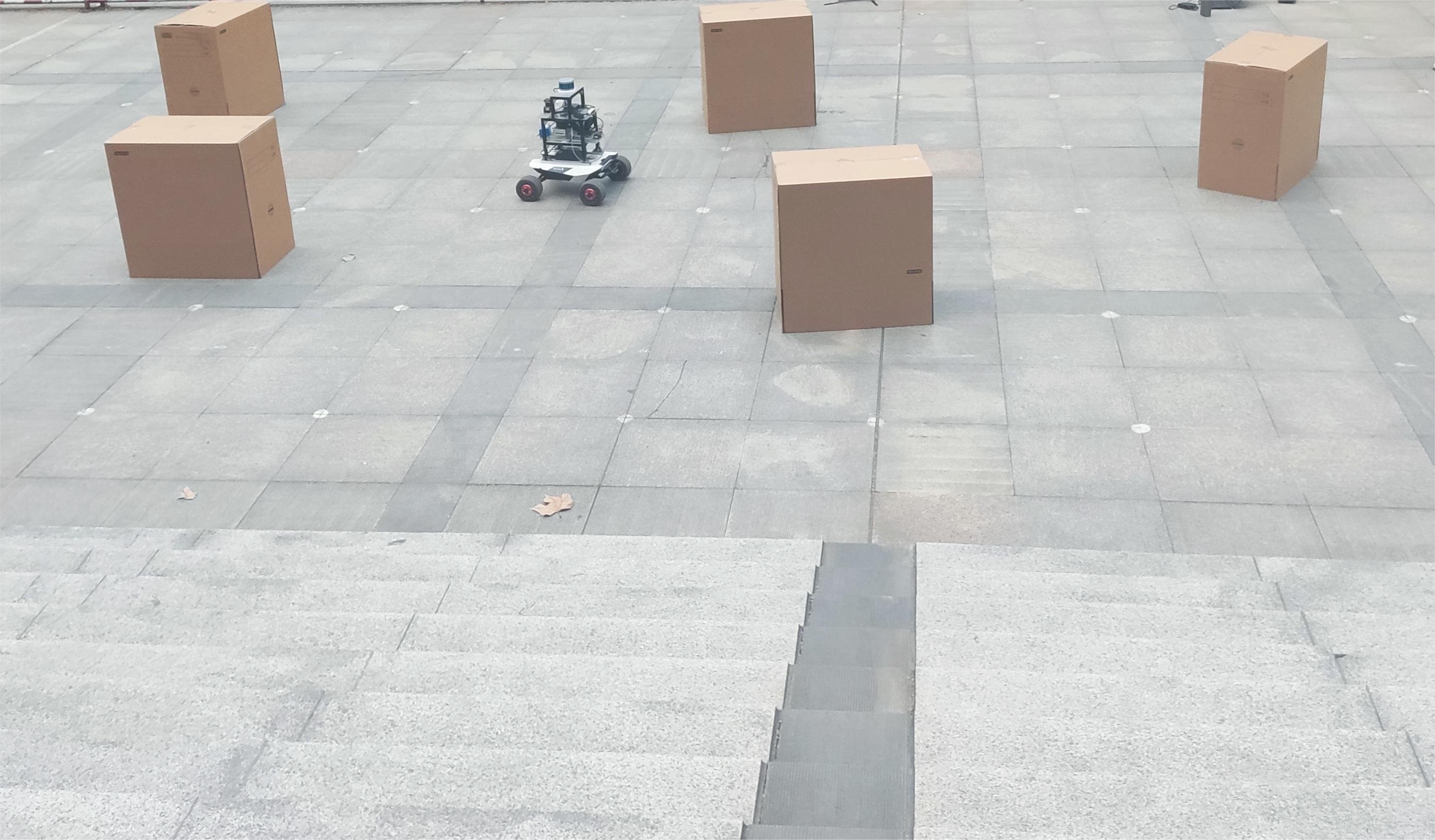}}
     \subfloat[\label{fig:PCM}]{
		\includegraphics[width=3.5cm]{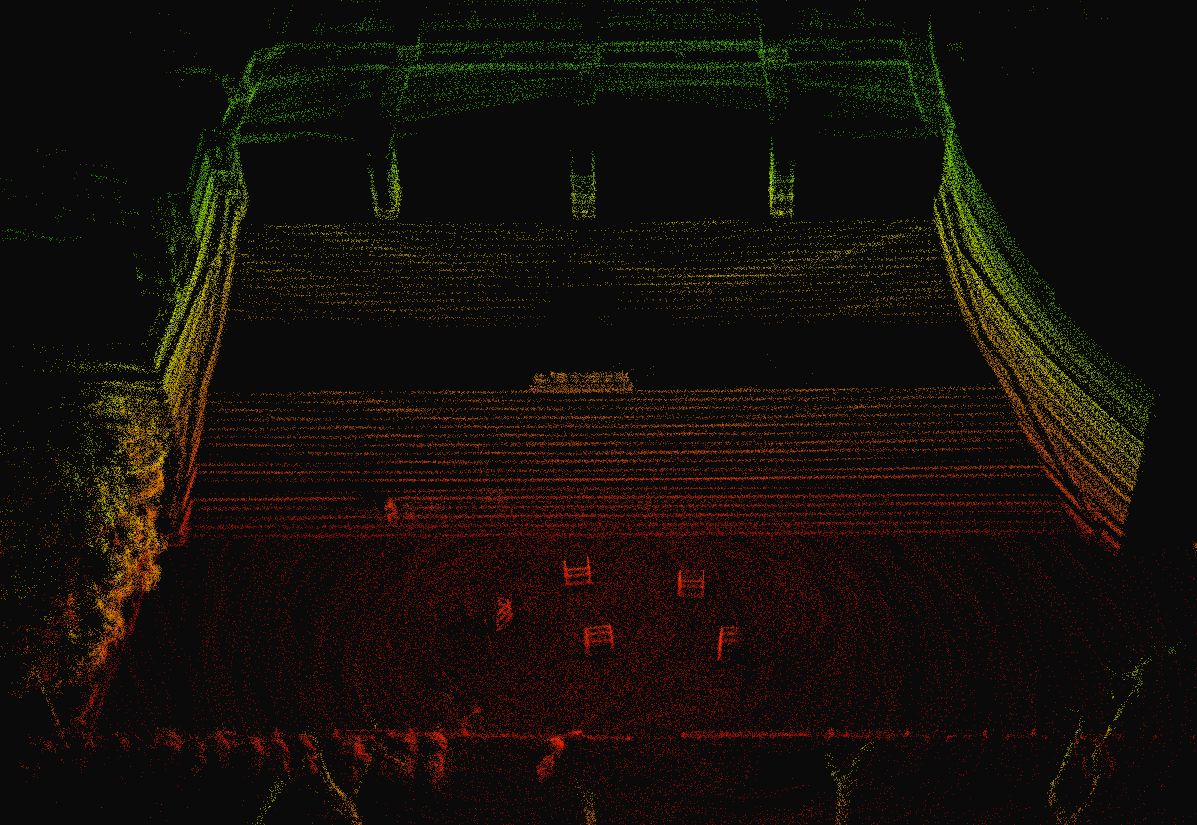}}
	\caption{The physical experiment scenario includes five obstacles. The size of each obstacle is $85\times 60 \times50 \text{mm}$. (a) actual scene, (b) constructed point cloud map.}
      \label{fig:Actual_Maps}
\end{figure}

\begin{figure}[htbp]
   \centering
      \setlength{\abovecaptionskip}{0.cm}
      \includegraphics[width=8cm]{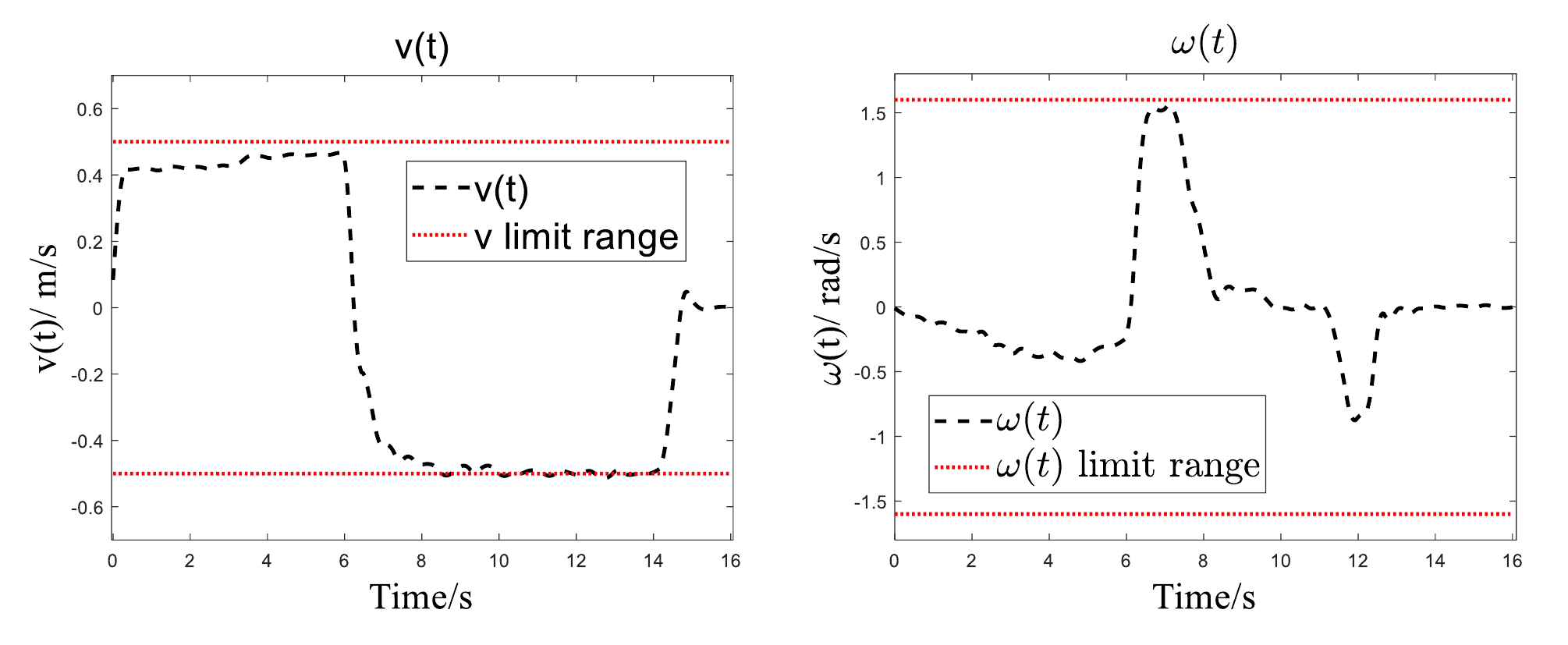}
      \caption{
      The corresponding linear velocity, linear acceleration, and angular velocity during SCOUT MINI's execution.}
      \label{fig:resPE}
\end{figure}

In the beginning, the map of the test environment was constructed using the Velodyne VLP-16 and the mapping method Lio-sam \cite{shan2020lio}. The test experiment occurred within a $\text{6 m} \times \text{8 m}$ map, where obstacles were strategically arranged using cardboard boxes. The decision to create a 3D map was driven by the necessity to account for obstacles of varying heights in the trajectory planning algorithm, ensuring safety. Fig.~\ref{fig:Actual_Maps} illustrates both the map of the test environment and the resulting 3D point cloud map. Throughout the physical experiments, we imposed maximum limits: the linear velocity ${v}^{\max}$ was capped at 0.5 m/s, while the angular velocity ${\omega}^{\max}$ was restricted to 1.6 rad/s. The experiment's results can be referenced in Fig.~\ref{fig:resPE}. For more information, please refer to the video submitted with our presentation.
\section{CONCLUSIONS}
This paper introduces a practical online fast trajectory planning framework for AGVs operating in obstacle-rich environments. The trajectory planning problem is formulated as an optimal control problem. Initially, the Modified Visibility Graph is used for rapid path planning and deriving path points. Subsequently, the Fast Safe Rectangular Corridor algorithm is employed to establish safe corridors quickly. This constrains AGVs' path points within corresponding rectangular boxes for optimization, effectively converting large-scale non-convex redundant obstacle avoidance constraints into linear and easily solvable box constraints. The effectiveness and superiority of our approach are validated through simulation and physical experiments.
%In future work, the methodology will be extended to multi-AGV systems, enabling the dynamic online replanning of each AGV's trajectory to address obstacles in unknown and dynamic environments.
%%%%%%%%%%%%%%%%%%%%%%%%%%%%%%%%%%%%%%%%%%%%%%%%%%%%%%%%%%%%%%%%%%%%%%%%%%%%%%%%
%\clearpage
\bibliographystyle{IEEEtran}	
\bibliography{scc}
\end{document}